# Real-Time Facial Expression Recognition using Facial Landmarks and Neural Networks

Mohammad A. Haghpanah
Human and Robot
Interaction Lab, School of
Electrical and Computer
Engineering,
University of Tehran
Tehran, Iran
amin.haghpanah@ut.ac.ir

Ehsan Saeedizade
Department of Computer Science
and Engineering,
University of Nevada,
Reno, USA
ehsansaeedizade@nevada.unr.edu

Mehdi Tale Masouleh
Human and Robot
Interaction Lab, School of
Electrical and Computer
Engineering,
University of Tehran
Tehran, Iran
m.t.masouleh@ut.ac.ir

Ahmad Kalhor
Human and Robot
Interaction Lab, School of
Electrical and Computer
Engineering,
University of Tehran
Tehran, Iran
akalhor@ut.ac.ir

*Abstract* – **This paper presents a lightweight algorithm for feature extraction, classification of seven different emotions, and facial expression recognition in a real-time manner based on static images of the human face. In this regard, a Multi-Layer Perceptron (MLP) neural network is trained based on the foregoing algorithm. In order to classify human faces, first, some pre-processing is applied to the input image, which can localize and cut out faces from it. In the next step, a facial landmark detection library is used, which can detect the landmarks of each face. Then, the human face is split into upper and lower faces, which enables the extraction of the desired features from each part. In the proposed model, both geometric and texture-based feature types are taken into account. After the feature extraction phase, a normalized vector of features is created. A 3-layer MLP is trained using these feature vectors, leading to 96% accuracy on the test set.**

*Index Terms* – Facial Expression Recognition, Emotion Recognition, Image Processing, Machine Vision, Machine Learning, Neural Network.

## 1. INTRODUCTION

Facial expression is a powerful and fast way for humans in order to communicate in their daily life and express their emotions. Today automatic or real-time Facial Expression Recognition (FER) has many applications in different areas such as human-computer interaction, virtual reality, human emotion analysis, cognitive science. In order to develop a reliable FER system, different kinds of expressions should be identified which seek to be recognized. In [1], it has been revealed that there are certain kinds of expressions which have universal, equal meaning among different cultures. Referring to the latter studies, it can be inferred that these expressions fall into six groups, namely, anger, disgust, fear, happiness, sadness, and surprise. In addition to these six categories, some systems also considered neutral as the seventh type of emotion [2]. Several systems are developed in order to describe different states of a face. The Facial Action

Coding System (FACS), among others, can be regarded as one of the most well-known systems. The FACS describes different changes in the human face in the form of 33 Action Units (AU), and it is possible to model almost all different facial expressions by using this system [3], [4].

Generally, there are two approaches to develop a FER system. As the first approach, some systems use a sequence of images captured from a neutral face to the peak state of emotions. In turn, some approaches are based on a single image of the face in order to recognize related emotion, and since they have access to less information, they usually result in less accuracy compared to the foremost [5], [6]. Besides the approach type for which a FER system is modeled, another classification is based on the type of features which are used in the process of recognition where a FER system can use one or both of these feature types. The first group of features is extracted based on the posture of organs in the face and the texture of the skin. The second type is geometric features which contain information about different positions and points on the face either in the processing of a static image or a sequence of images and use the movement of the positions and points among a sequence. In order to extract geometric features, one way is to use facial landmarks. Landmarks are key points of the face, and therefore they are valuable for facial analysis. Several researches have been performed on the subject of facial landmark detection, but they are out of the scope of this paper. In this work, a python library called dlib is used in order to detect these points [2], [6], [7].

FER systems use different techniques for classification such as Multi-Layer Perceptron neural networks (MLP), Support Vector Machine (SVM), Hidden Markov Model (HMM), Convolutional Neural Network (CNN) [8], [6], [9], [10]. For the purposes of this paper, both geometric and texture-based features are used in order to train an MLP neural network. The main goal of this paper consists in categorizing single images of a face into seven different expressions. Moreover, it has been tempted to extract simple but precise features for the FER system. Therefore, the proposed algorithm would have a simultaneously high speed in processing and high accuracy, which is a definite asset in practice. The latter issues make the proposed algorithm suitable for real-time recognition.

The remainder of this paper is organized as follows. In Section 2, a review of some works related to the FER subject is provided. Details of the datasets used in this work are described in Section 3. Section 4 describes in detail the proposed algorithms and architecture of the corresponding neural network. Section 5 is devoted to evaluating the proposed model and its accuracy under different conditions. Finally, the paper concludes with some hints as ongoing works.

## 2. RELATED WORKS

Arushi and Vivek in [10] implemented multiple CNNs to classify static face images into seven categories. They used Kaggle's dataset to train their networks. In the latter study, three different classifiers are designed and implemented; namely, a baseline with one convolutional layer, a five-layer CNN and a deeper parametrized CNN. The objective of the foregoing study consisted in fine-tuning different parameters of these models by using VGG16 and VGGFace models. From the results of the latter, an accuracy of 48% in Facial Expression Recognition is reported.

In [11], the feasibility of using CNNs for facial expression recognition in real-time is examined in which the network is fed by a stream of images or video. According to the provided report, they achieved an overall 90% and 57.1% accuracy on train and test datasets, respectively. In the latter study, an extended Cohn-Kanade (CK+) and Japanese Female Facial Expression (JAFFE) datasets have been used [12], [13]. In addition to these datasets, a customized dataset is gathered, and a new dataset is created in order to improve the accuracy of the network. The VGG_S network proposed in [14] was used to the end of improving its accuracy. In the latter paper, it has been also stated that due to resource limitations, only some of the layers from the original networks were trained [11].

Ying-li Takeo, and Jeffery F. in [15], developed a system to recognize AU to be used in a FER framework. For this purpose, first, different geometric features from the upper and lower part of a face from lips, eyes and brows are extracted. These permanent features, in addition to transient ones, are fed into two different neural networks where each one predicts existed AUs in upper or lower parts. In the proposed approach, CK and Ekman-Hager Facial Action Exemplars are used as their dataset, which leads to an average of 96.4% and 96.7% accuracy in AUs recognition in the upper and lower face, respectively. Although the proposed approach for the feature extraction phase is similar to the present work, a FER system has not been developed and has a different objective. The approach presented in this paper is different from the one presented in [15] since extracted features are directly used as the input of a single neural network which leads to recognizing the top most related expression to a human face's image.

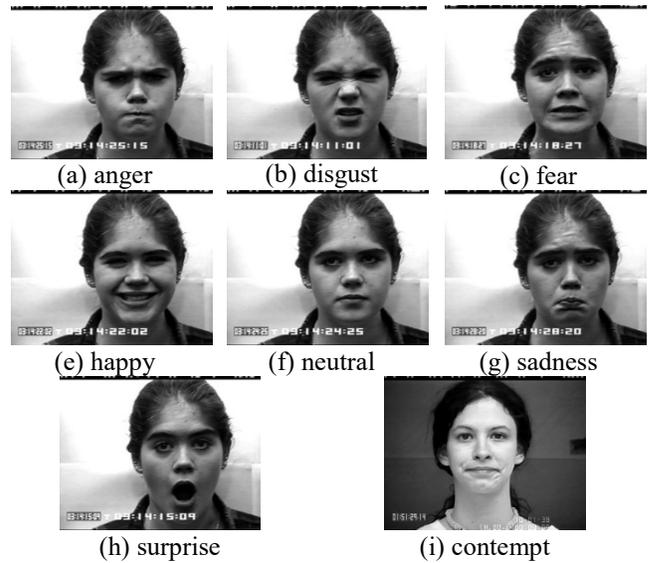

Figure 1. Different emotions of two subjects from CK+ dataset [12].

Table 1. Number of selected images for each emotion from CK+ dataset [12].

| Emotion | Number of Images |
|---------|------------------|
| Anger | 45 |
| Disgust | 59 |
| Fear | 25 |
| Happy | 69 |
| Neutral | 593 |
| Sadness | 28 |
| Surprise | 83 |

However, it should be noted that the study conducted in [16] is similar to the one presented in this paper in some aspects. In fact, in the foregoing study an MLP network is used to classify static images into six known universal emotions plus neutral and the author reports an accuracy of 84.9-98.2% on different categories of emotions. Feature extraction is performed after a pre-processing phase and landmark detection which extracts the region of interest from the face. The features are basically the Euclidean distance between extracted feature points. Although the general flow of the processing in [16] is similar to the one presented in this paper, the network is different as it is trained with images from Karolinska Directed Emotional Faces (KDEF) dataset. Another key difference can be regarded as the difference between the features which are used for training the network. While features in [16] are based on Euclidean distance, in this paper, both geometric features based on angles between the landmark points and texture features from the skin are used.

## 3. DATASET

In order to train and test the presented model, firstly, the well-known Extended Cohn-Kanade (CK+) dataset, which is released with the purpose of emotion recognition, is used. The CK+ dataset includes well-defined expressions in 593 sequences from neutral frame (first frame) to peak expression (last frame) of 123 subjects in 8 different expressions. In this study, the first frame is selected as neutral and the last one for peak emotion [17] [12]. Also, images of contempt emotion are removed because there are only 18 of them in the dataset. Figure 1 shows the different emotions of two subjects in this dataset.

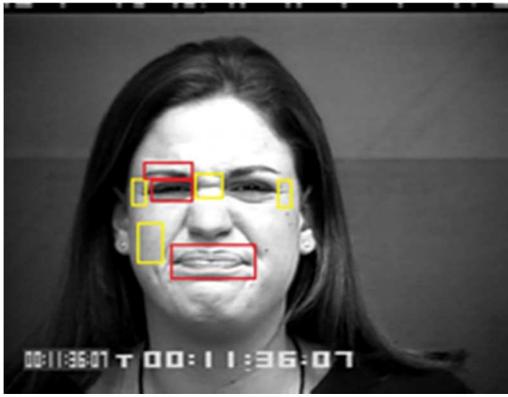

*Figure 2. Regions of interest in the proposed algorithm. The face image is taken from [12].*

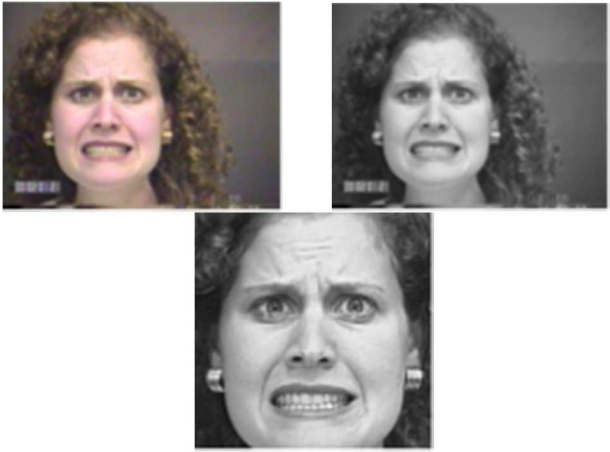

Figure 3. *Result of applying the pre-processing and face detection steps on an image from CK+ dataset [12].*

A total number of 902 images are selected for the training and test phase. 722 of all images are used to train the network, and the rest of the 180 images are used for the test phase. Table 1 shows the distribution of images over each emotion.

## 4. PROPOSED ALGORITHM
### AND
### IMPLEMENTATION DETAILS

In this section, details of each step of the proposed algorithm are provided. Briefly, after a simple pre-processing on the input image, facial landmarks and edges on the face are detected and the feature vector is extracted for all images in the dataset. Then, a network will be trained by using this information.

### 4.1 REGION OF INTEREST

By observing the different emotions in the two datasets, it can be deduced that these emotions have a clear definition. As a result, the Regions of Interest (ROI) in one's face are defined as the right eyelid and eyebrow, lips, corner of eyes, nose and cheeks. According to these definitions, 10 different geometric and texture-based features are recognized to train a network. In fact, five of them are geometric characteristics with the type of angle. In Section 4.2.4, more details about each feature are provided. Figure 2 shows the ROI used in this work. The red rectangles are regions from which the geometric features are extracted. Texture-based features are similarly extracted from yellow regions.

### 4.2. PROPOSED ALGORITHM

In order to discuss the proposed algorithm, the whole approach is divided into 4 steps. The details of each step are provided in the following.

#### 4.2.1. PRE-PROCESSING

The purpose of pre-processing phase consists in preparing input images for the feature extraction phase. Because images in the datasets are both grayed and colorful, the first step is to convert all images into grayscale. This step is done using OpenCV [18] library, which is an open-source computer vision and machine learning software library mainly aimed at real-time computer vision.

#### 4.2.2. FACE DETECTION

After converting images into grayscale, it is necessary to eliminate any extra part of them except faces. Dlib [7] library is used for the face detection phase. By doing so, better accuracy in the landmark detection step can be achieved. Figure 3 shows the resulting image after applying steps 1 and 2 on an input image.

#### 4.2.3. LANDMARK DETECTION

There are multiple types of facial landmarks such as 5-points and 68-points [19]. In this study, 68-points facial landmark is used, which gives adequate points for the feature extraction part. Furthermore, dlib [7] has a real-time and accurate implementation which makes it possible for the proposed algorithm to be both real-time and precise simultaneously. Accurately estimating landmark positions is very important as the feature extraction part completely relies on this step, and errors in localization will result in a non-precise feature vector [2].

#### 4.2.4.1. GEOMETRIC FEATURES

After detecting the face's key points, their rotations around the roll and yaw axis are normalized, respectively. This normalization helps the neural network to fit better because the network no longer needs to learn anything about the face rotations around different axes. Moreover, it improves the ability of the proposed method to generalize and predict the emotions of new faces which are not in the dataset.

In order to normalize key points around the roll axis, first, the center point of each eye is found. Then the center of these two points is considered as the origin point. Therefore, the angle (α) between the line which connects the center of the eyes and the horizontal line which passes through the origin point can be calculated. Finally, as pectited schematicallty in Fig. 4, a rotation is applied to the key points around the origin point by α, and then, points are normalized around the roll axis. Next, the result points will be normalized around the yaw axis. Each point of these 68 points from the left side of a face has a symmetry point on the right side because a face is simply symmetrical around the vertical axis.

The proposed approach for the yaw normalization is to average all points with their corresponding symmetry point. As the key points are normalized around the roll and yaw axes, 5 geometric feature vectors from 5 regions are extracted, right eyelid and eyebrow, upper upper lip, lower upper lip, upper lower lip, and lower lower lip. Figure 5 shows these feature vectors separately. The angle between key points of each region is calculated as its feature vector. In some studies, Euclidean distance between these points is calculated but, in this study, angle is used instead of distance which makes the proposed algorithm scale-invariant. Thus, different sizes of faces or images will not affect the feature vectors. This is a magnificent advantage of the proposed feature vectors, which increases the ability of the network to generalize for images which are not in the dataset. Moreover, the angle's value is in the range [-π, +π], so it is near [-3.14, +3.14], and it does not need to be normalized before feeding to neural networks.

#### 4.2.4.2. TEXTURE-BASED FEATURES

In some emotions, such as disgust and anger, there are some wrinkles around the eyes and cheeks. The latter issue is useful and can help to distinguish different emotions. To this end, five regions are chosen, between two eyes, right side of the right eye, left side of the left eye, right cheek, left cheek. In order to detect wrinkles, edge detection algorithms are used, especially the horizontal Sobel edge detector [20]. As shown in Fig. 6, this algorithm puts some white pixels where wrinkles are located. The value of each pixel is a number in the range 0-255, specifying the edge intensity on that pixel. There will be no white pixels if there are no wrinkles in a region. Finally, for each region, the density of wrinkles in that region is calculated as its feature value as follows:

$$\text{Density} = \frac{1}{N} \times \sum_{i=1}^{N} \frac{\text{region\_pixels}[i]}{255} \qquad (1)$$

The above equation shows how this density is calculated, where $N$ indicates the number of pixels in a region, *region pixels* indicates the value of the pixels of that region (which is in range 0-255), and the *density* is the average of white pixels in the region (which is in range 0-1). Similar to the geometric features, these features are also in the range [0, 1], and there is no need to normalize them too while passing to the neural network.

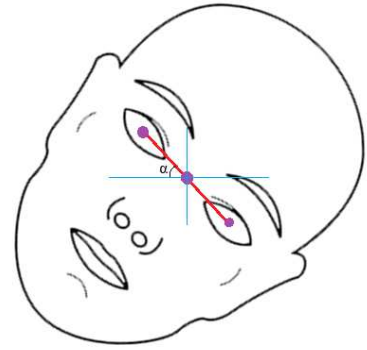

*Figure 4. Normalizing the face around roll axis.*

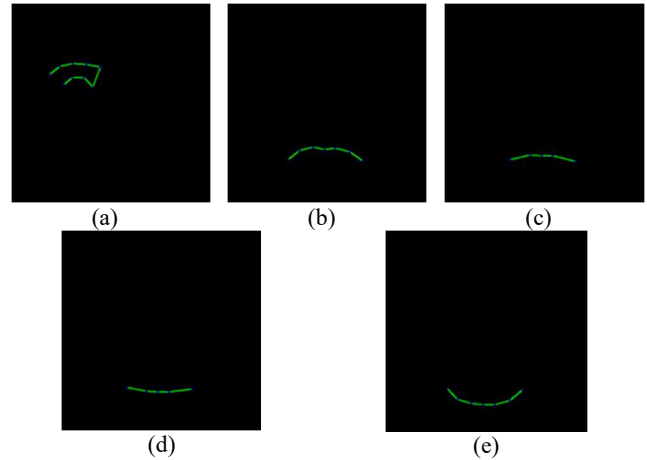

*Figure 5. Five regions from which feature vectors are extracted.*

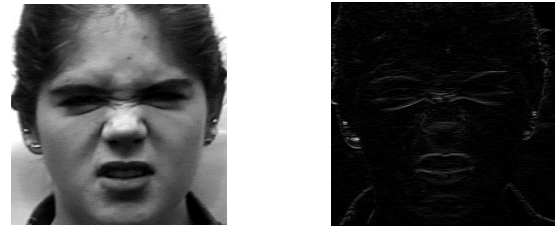

*Figure 6. Sobel edge detection result on a face from CK+ dataset [12].*

Figure 7 shows three of the five mentioned ROIs. The left images show the resulting image from Fig. 6, in which the pixels' color outside of that ROI is blackened. The right images show ROIs, where the background color of the ROI is white, and the edge intensity values are specified by black pixels (255 – RealEdgeIntensity).

Hence all feature values lie in a small range [-3.14, +3.14], they are essentially normalized, and there is no need to normalize them separately using other methods, such as

Standard Scaling and Min-Max Scaling for feeding to the neural network.

### 4.3. NETWORK ARCHITECTURE

So far, some feature vectors have been extracted from a raw image of a human face. The last step consists in classifying different emotions using these features. To do so, a classifier should be acquired. There are multiple types of classifiers, such as K-Nearest Neighbours (KNN), Support Vector Machines (SVM), Multi-Layer Perceptron (MLP), to be named but a few.

In this study, an MLP is used because its ability in classification tasks in the past years is proven. For each face, a feature vector is calculated according to the previous section. This feature vector is essentially normalized and can be fed into the network without any processing. Table 2 shows the details of each layer in our network. An MLP with two hidden layers is used, and the output layer is softmax. Batch normalization and Dropout techniques are applied to each hidden layer in order to prevent the network from overfitting.

### 5. EXPERIMENTAL ANALYSIS

In this section, the proposed model is evaluated on the CK+ dataset. The accuracy of the model for each emotion and the whole dataset is presented. Furthermore, some situations in which the model fails to predict properly are discussed.

### 5.1. GENERAL REPORT

In order to train and evaluate the proposed method, K-Fold cross-validation with 5 folds is used. Usually, when the dataset is large, and the network is complicated, the comptuatioanl time for training and evaluating stages will increase, which makes impossible to use K-Fold. But, in this method, the model can be evaluated with K-Fold cross-validation because not only that for each sample in the dataset the feature vectors are calculated and stored, but also the network is a simple MLP and, it can be trained on the data relatively fast.

Table 3 shows the accuracy on the train, test and whole dataset for each fold, as well as the total accuracy with its 95% confidence interval. As it can be seen, the accuracy of 96.12±2.56% on the test set, 98.94±0.80% on the training set, and 98.38±0.47% on all samples in the dataset are achieved.

Table 4 shows the confusion matrix of the fold with the highest accuracy on the test set. This matrix indicates that all emotions except sadness are predicted with an accuracy of +85%, while accuracy for sadness is 66%. This is a great gap, and it will be discussed in Section 5.2.

### 5.2. FAULT REPORT

In Section 4.2.3, it has been noticed that any error in estimating key points will affect, as a consequence, the feature extraction part and classification. Figure 8 shows the extracted key points for some faces with sadness emotion. As it can be seen, there is a significant error in estimating lip points. Therefore, this error affects the features extracted by the proposed algorithm, and as a result, the classifier cannot classify this emotion properly.

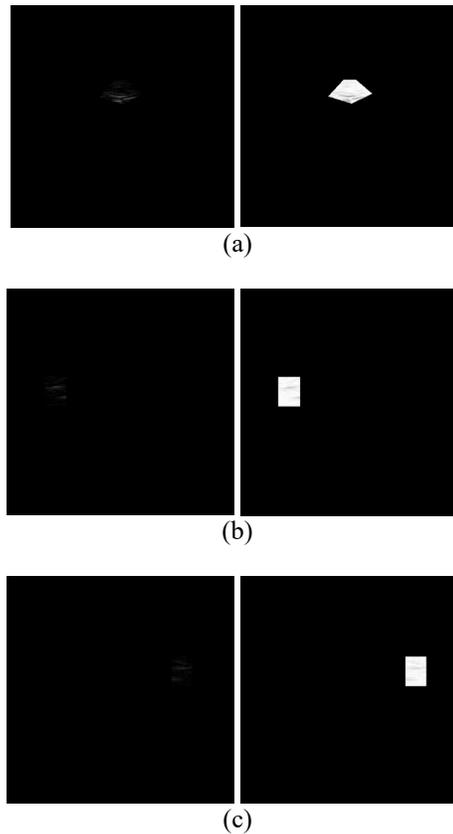

Figure 7. Edges used in texture-based features. (a). between eyes (b). right side of the right eye. (c). left side of the left eye.

Table 2. Details of each layer in the proposed network.

| Layer Number | Layer Name | Parameters |
|---|---|---|
| 1 | Dense | N=1024 |
| 2 | Batch Norm. | Momentum=0.99 |
| 3 | Activation | Leaky Relu |
| 4 | Dropout | Rate=0.3 |
| 5 | Dense | N=1024 |
| 6 | Batch Norm. | Momentum=0.99 |
| 7 | Activation | Leaky Relu |
| 8 | Dropout | Rate=0.3 |
| 9 | Dense | N= # of classes |
| 10 | Activation | Softmax |

Table 3. Accuracy on train, test, and whole dataset for different folds.

| Data | Fold-1 | Fold-2 | Fold-3 | Fold-4 | Fold-5 | Total |
|---|---|---|---|---|---|---|
| Train | 98.61 | 99.72 | 98.89 | 98.89 | 98.61 | 98.94±0.80 |
| Test | 96.69 | 93.92 | 97.78 | 96.67 | 95.56 | 96.12±2.56 |
| All | 98.23 | 98.56 | 98.67 | 98.45 | 98.00 | 98.38±0.47 |

### 5.3. REAL-TIME PERFORMANCE

The proposed algorithm has three main time-consuming phases, namely, landmark detection, feature extraction, and prediction using MLP. For an input image, the processing time of each mentioned phase and the whole algorithm is presented in Table 5. The test was performed on a single core of an Intel Core i7-4710HQ processor, using a 480p laptop webcam. It can be inferred that the algorithm is fast enough to be executed in real-time applications, and it achieves an FPS of 20 on average.

## 5.4. COMPARISON WITH OTHER METHODS

In this section, the proposed algorithm is compared with multiple state-of-the-arts FER methods and researches. It is worth mentioning that it may not be fair to compare different FER methods as they leverage other learning techniques and datasets to train their models. Moreover, an attempt was made to select studies that used similar datasets as the proposed method. In the proposed method, an MLP neural network is trained to predict seven different expressions by using the CK+ dataset with 96% accuracy. Kashif et al. achieved an accuracy of 99.4% on the same dataset by using SVM, KNN and NP machine learning algorithms [21]. However, they consider six different expressions in their work. Gupta achieves 92.1% accuracy on real-time classifying eight different expressions using SVM [22]. Khan leverages MLP neural network and facial landmarks to classify seven different expressions [16]. However, it differs from the proposed method since it is trained on a different dataset, and its accuracy ranges from 84 to 95%. Duncan et al. shows the possibility of using CNN for real-time FER. The authors achieved 57.1% accuracy for the classification of 7 expressions on the test set [23]. Qiu and Wan also proposed MLP based model and leverage landmarks to predict seven emotions. However, their model's accuracy is 92% on the CK+ dataset [24]. Table 6 shows the summary of the comparison of the proposed method with other FER methods.

## 6. CONCLUSION AND FUTURE WORKS

In this paper, a new approach was proposed for the Real-Time Facial Expression Recognition problem with seven emotions. An accuracy of 96% on the test set was achieved, which is comparative to the aforementioned state-of-the-art algorithms mentioned in. Although the accuracy achieved by this study is 2-3% lower than the CNN-based algorithms in the literature, the speed is considerably high. The proposed algorithm is much faster in both feature extraction and classification parts in comparison with CNN-based methods. Thus, it can be used in real-time manners. Both geometric and texture-based features were taken into account in order to increase accuracy. An MLP network processes these features and classifies the emotion of an input face. A great improvement in performance is achieved by the pre-processing and normalization steps, which increased the ability of the proposed method in predicting non-seen images. More features can be computed for distinguishing different emotions and improve performance as a future work. In the facial keypoints extraction part, dlib was used because of its high speed and accuracy. However, in some emotions such as sadness, this library has a significant error in predicting the true points, which leads to a false classification. In the future, some post-processing methods can be considered in order to reduce the error and improve the results. Besides the roll and yaw normalization stages, also pitch normalization is under investigation as ongoing work, and it can be applied to the landmarks in order to reduce the effect of face rotations in emotion prediction.

*Table 4. The confusion matrix of the fold with the highest accuracy on test set.*

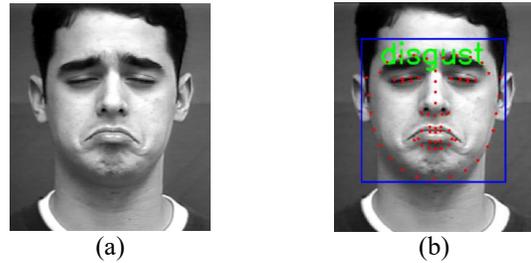

|       | AN    | DI    | FE     | HA     | NE     | SA    | SU     |
|-------|-------|-------|--------|--------|--------|-------|--------|
| AN    | 87.50 | 0.00  | 0.00   | 0.00   | 12.50  | 0.00  | 0.00   |
| DI    | 0.00  | 92.31 | 0.00   | 0.00   | 7.69   | 0.00  | 0.00   |
| FE    | 0.00  | 0.00  | 100.00 | 0.00   | 0.00   | 0.00  | 0.00   |
| HA    | 0.00  | 0.00  | 0.00   | 100.00 | 0.00   | 0.00  | 0.00   |
| NE    | 0.00  | 0.00  | 0.00   | 0.00   | 100.00 | 0.00  | 0.00   |
| SA    | 0.00  | 0.00  | 0.00   | 0.00   | 33.33  | 66.67 | 0.00   |
| SU    | 0.00  | 0.00  | 0.00   | 0.00   | 0.00   | 0.00  | 100.00 |

*Figure 8. Landmarks and predicted emotion of a sad face. Image is taken from [12] and the colored results are from the algorithm proposed in this paper.*

*Table 5. Processing time of each phase of the proposed method.*

| Phase | Landmark Detection | Feature Extraction | Prediction using MLP | Total |
|-------|--------------------|--------------------|----------------------|-------|
| **Processing Time (s)** | 0.047 | 0.002 | 0.001 | **0.05** |

*Table 6. Comparisopn of the proposed method with other methods.*

| Ref | Method | Dataset | # of Classes | Real-time | Accuracy |
|-----|--------|---------|--------------|-----------|----------|
| [21] | SVM / KNN / NB+LBP | CK+ / Oulu–CASIA | 6 | | 99.4% |
| [22] | SVM | CK+ | 8 | * | 92.1% |
| [16] | MLP | KDEF | 7 | | 84-95% |
| [23] | CNN | CK+ / JAFFE / their own DS | 7 | * | 57.1% |
| [24] | MLP | CK+ | 7 | | 92% |
| **Proposed Method** | | CK+ | 7 | * | **96.12%** |